\documentclass[runningheads]{llncs}

\usepackage[T1]{fontenc}
\usepackage{graphicx}
\usepackage{marvosym}
\usepackage{xcolor}
\usepackage{tikz}
\usepackage{booktabs}
\usepackage{float}
\usepackage{placeins}
\usepackage{url}
\usepackage{multirow}
\usepackage[table]{xcolor}
\usepackage{amssymb}
\usepackage{amsmath}

\usetikzlibrary{arrows.meta,positioning,shapes.geometric}
\definecolor{inputblue}{HTML}{D9EAF7}
\definecolor{rerankpurple}{HTML}{E9D8FD}
\definecolor{evidencegreen}{HTML}{DFF3E4}
\definecolor{classifierorange}{HTML}{FFE6C7}
\definecolor{outputred}{HTML}{F9D6D5}
\definecolor{linegray}{HTML}{4A5568}

\begin{document}
\title{HNR-DAC: Hard-Negative Reranking and Distribution-Aligned Classification for Scientific Claim Verification}
\titlerunning{HNR-DAC}

\author{Zhenchao Wang\inst{1,2} \and
Xin Chen\inst{2} \and 
Luoxi Zhang\inst{2} \and
Min Yang\inst{2,3} \and
Shiwen Ni\inst{2}\textsuperscript{(\Letter)}}

\authorrunning{Z. Wang et al.}

\institute{
Southern University of Science and Technology, Shenzhen, China\\
\email{
12533253@mail.sustech.edu.cn}  \and
Institute of Artificial Intelligence, Shenzhen University of Advanced Technology, Shenzhen, China\\
\email{250701451@stu.suat-sz.edu.cn, \{zhangluoxi,nishiwen\}@suat-sz.edu.cn}\and
Shenzhen Institutes of Advanced Technology, Chinese Academy of Sciences, Shenzhen, China\\
\email{min.yang@siat.ac.cn}}

\maketitle

\begin{abstract}
Scientific claim verification over a cited paper requires predicting the
claim--paper relation and identifying the paragraphs that justify that
prediction. This setting poses two linked challenges: within-paper
distractors often resemble genuine evidence, while a classifier trained on gold evidence must operate on retrieved evidence at inference. We present
HNR-DAC, a two-stage framework that trains each stage on the cases it will
actually encounter. Hard-Negative Reranking (HNR) quantifies 
evidence confusability using a base reranker's scores on non-gold
paragraphs and contrasts gold evidence against the most confusable
candidates. Distribution-Aligned Classification (DAC) trains on the Top-1
paragraph produced by the same frozen HNR used to construct inference
inputs, while HNR's Top-3 paragraph identifiers provide the evidence output.
On the NLPCC~2026 Task~10 Track~2, the final configuration
obtains 97.21\% Hit@3, 95.79\% Macro-F1, 94.47\% Joint@3, and an average score of 95.13\%. The corresponding submission ranks third on the official
Track~2 leaderboard while achieving the highest overall Macro-F1 of
93.05\%, alongside 70.16\% Joint@3 and an average score of 81.61\%.
\end{abstract}

\section{Introduction}

Large language models (LLMs) can generate literature-review drafts and
citation-supported scientific text~\cite{chen-etal-2025-repreguard,gao-etal-2023-enabling,wu-etal-2025-survey}, but their claims may remain
unsupported by or inaccurately represent the cited
sources~\cite{walters2023fabrication,wu2025sourcecheckup}. The reliability of AI-assisted
scientific reporting therefore depends not only on fluent generation, but
also on whether a claim remains faithful to the evidence it cites.
NLPCC~2026 Shared Task~10 is designed around this
problem~\cite{nlpcc2026task10}. This paper focuses on Track~2, where each
instance provides an atomic AI-generated claim and the structured full text
of its cited paper. A system must determine whether the paper supports,
overstates, is only topically related to, or is irrelevant to the claim,
while also returning ranked evidence paragraph identifiers.

The task follows the retrieve-and-verify paradigm established by FEVER and
SciFact~\cite{thorne2018fever,wadden2020scifact}, but considers a more
constrained retrieval setting: the cited paper is already fixed, evidence
must be identified at paragraph level, and the relation label depends on
whether the selected paragraphs support the exact scope of the claim.

This fixed-paper setting creates two stage-specific difficulties. For
evidence reranking, topical overlap is pervasive within the cited paper:
many non-gold paragraphs share the claim's entities, terminology, and
experimental setting. The most informative negatives are therefore not
arbitrary non-gold paragraphs, but those that the current reranker itself
treats as plausible evidence. We refer to this claim-specific and
model-conditioned tendency as evidence confusability. For relation classification, training only on gold evidence exposes the classifier to cleaner contexts than the automatically retrieved paragraphs available at inference. The reranker therefore requires evidence-confusable
within-paper negatives, whereas the classifier requires contexts produced
by the retrieval pipeline it will encounter at inference.

We propose HNR-DAC, a two-stage framework that trains each stage on the
cases it will actually encounter. Hard-Negative Reranking (HNR)
operationalizes evidence confusability using the scores assigned by a base
reranker to non-gold paragraphs. It then trains a cross-encoder to
distinguish gold evidence paragraphs from the most confusable candidates using a
group-wise contrastive objective. Distribution-Aligned Classification (DAC)
freezes the trained HNR and fine-tunes a four-way classifier on its
top-ranked paragraph, aligning the retrieval process used to construct
classifier inputs across training and inference. This construction applies
to all four classes, including non-empty Top-1 inputs for
\textsc{Irrelevant} claims. At inference, the DAC relation label is paired
with HNR's Top-3 paragraph identifiers for evidence output. On the
development set, the final configuration reaches 97.21\% Hit@3, 95.79\%
Macro-F1, 94.47\% Joint@3, and a 95.13\% Average Score, and the
corresponding submission ranks third on the official Track~2
leaderboard~\cite{nlpcc2026task10leaderboard}.

The main contributions of this work are as follows:
\begin{itemize}
    \item We introduce HNR, which quantifies evidence confusability using
    base-reranker scores and trains against the most confusable non-gold
    paragraphs within each cited paper.

    \item We present DAC, which trains the relation classifier on the Top-1
    paragraph produced by the same frozen HNR used at inference, while
    retaining the Top-3 paragraph identifiers for evidence output.

    \item We demonstrate that confusability-guided negatives consistently
    outperform random negatives and that HNR Top-1 provides the strongest
    classifier context; the full system ranks third on the official
    leaderboard with the highest overall Macro-F1.
\end{itemize}

\section{Related Work}

\subsection{Citation Faithfulness and Scientific Claim Verification}

Citation faithfulness asks whether a generated statement can be traced to an identified source and whether that source supports the statement~\cite{gao-etal-2023-enabling,rashkin-etal-2023-measuring}, while fine-grained citation evaluation shows that support is not purely binary~\cite{zhang-etal-2024-towards-fine-grained}. Thorne et al. introduced FEVER for evidence-based fact verification over textual sources, while Wadden et al. introduced SciFact for scientific claim verification and SciFact-Open for the open-domain setting~\cite{thorne2018fever,wadden2020scifact,wadden-etal-2022-scifact}. Unlike open-domain verification, NLPCC 2026 Task~10 Track~2 fixes the cited paper and requires both a four-way relation label and ranked evidence paragraph identifiers~\cite{nlpcc2026task10}. The main challenge therefore shifts from document retrieval to distinguishing topically related paragraphs from those that support the comparison or scope expressed by the claim.

\subsection{Evidence Reranking and Retrieval-Conditioned Classification}

Evidence reranking and relation classification are commonly combined in scientific claim verification. Karpukhin et al. proposed DPR, while Nogueira et al. used BERT and T5 for neural reranking~\cite{karpukhin2020dpr,nogueira2019passage,nogueira2020document}. Hard-negative training methods, including ANCE and RocketQA, use high-scoring incorrect candidates to improve retrieval~\cite{xiong2021ance,qu-etal-2021-rocketqa,zhan2021optimizing}. For scientific claim verification, VerT5erini and MultiVerS connect evidence selection and label prediction through a pipeline and joint prediction, respectively~\cite{pradeep-etal-2021-scientific,wadden-etal-2022-multivers}. In the fixed-paper setting, HNR mines high-scoring non-gold paragraphs as hard negatives to improve within-paper evidence discrimination. DAC constructs classifier training inputs from the outputs of the same frozen HNR used at inference, reducing the discrepancy between gold evidence at training and retrieved evidence at inference.
\FloatBarrier
\section{Methodology}

\begin{figure*}[t]
    \centering
    \includegraphics[width=0.95\textwidth, trim=0 0 0 0]{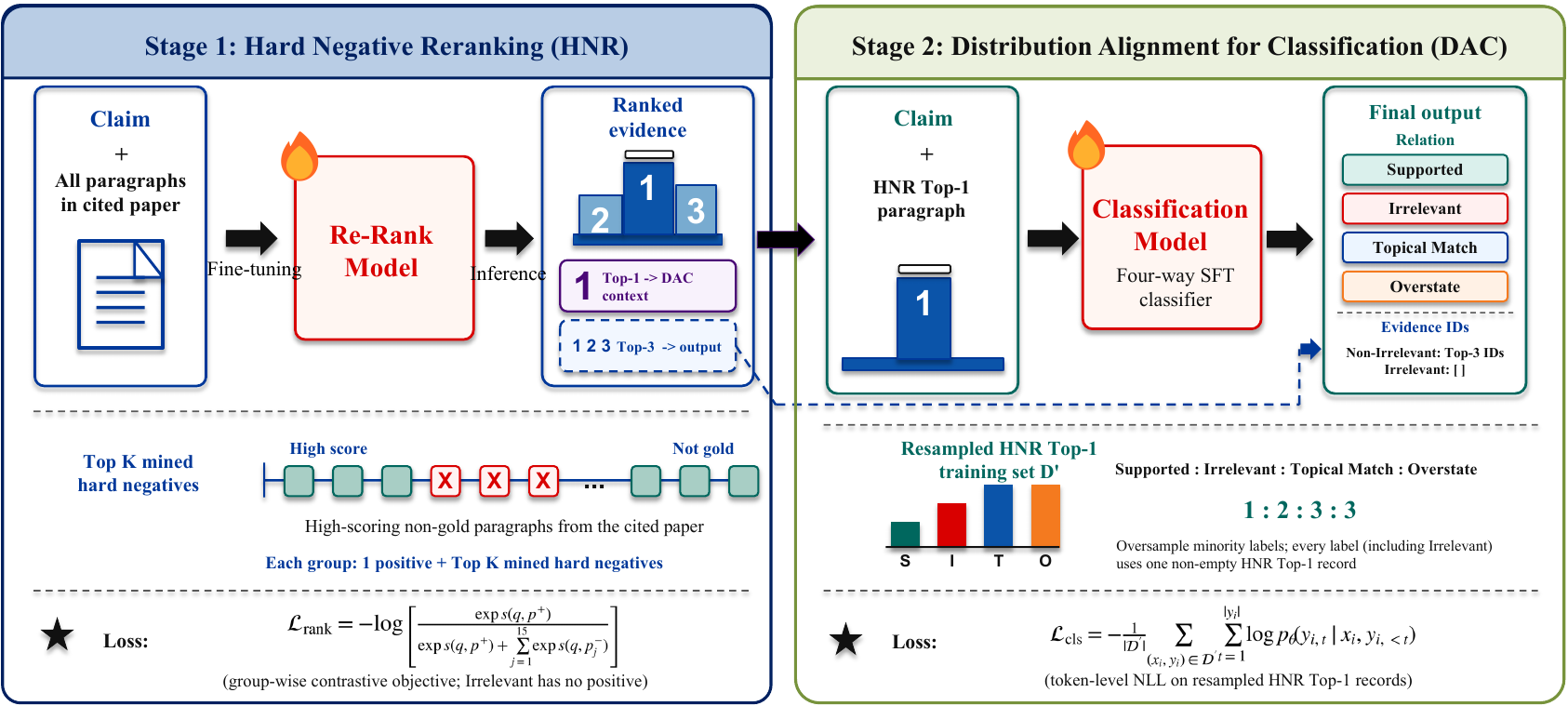}
    \caption{Overview of HNR-DAC. HNR mines evidence-confusable non-gold paragraphs for reranker training; its top-ranked paragraph conditions DAC, while the top-three paragraph identifiers bypass DAC for the final evidence output.}
    \label{fig:framework}
\end{figure*}

\subsection{Framework Overview}
\label{subsec:framework}

Track~2 considers claim verification over a fixed cited paper. Given an
atomic claim $q$ and a paper segmented into paragraphs
$P=\{p_i\}_{i=1}^{n}$, the system must predict a relation label
\[
y \in \mathcal{Y}
=
\{\textsc{Supported},\textsc{Irrelevant},
\textsc{Topical Match},\textsc{Overstate}\},
\]
together with an ordered list of up to three evidence paragraph identifiers. In the training data, each non-Irrelevant claim is paired with one or more human-annotated gold evidence paragraphs, whereas Irrelevant claims have no gold evidence annotations. These annotations are used only to construct training supervision and are unavailable at inference.

Although relation prediction and evidence identification are evaluated
jointly, the two stages face different learning conditions. Evidence
reranking must distinguish annotated evidence from evidence-confusable
distractors within the same paper, whereas relation classification must
judge the fine-grained claim--paper relation from automatically retrieved
context. Moreover, the context most useful for classification need not
contain the same number of paragraphs required for evidence output. These
differences motivate stage-specific training connected through an explicit
ranked-evidence interface.

As illustrated in Figure~\ref{fig:framework}, HNR-DAC consists of two
specialized stages. HNR assigns a relevance score to
each claim--paragraph pair and produces an ordered paragraph ranking.
DAC predicts the relation label from the
claim and HNR's highest-ranked paragraph,
\begin{equation}
\widehat{y}
=
f_{\mathrm{DAC}}
\bigl(q,\widehat{E}_1(q)\bigr),
\end{equation}
where $\widehat{E}_1(q)$ denotes the top-ranked paragraph produced by HNR. In
parallel, the top-three ranked paragraphs $\widehat{E}_3(q)$ bypass DAC and
are returned as the evidence output; following the task protocol, the evidence
list is set to empty when the predicted label is \textsc{Irrelevant}.

\subsection{Hard-Negative Reranking via Evidence Confusability}
\label{subsec:hnr}

Paragraphs within the same scientific paper often share entities,
terminology, methods, and experimental settings. Consequently, a paragraph
may appear highly relevant to a claim while failing to support its specific
finding, comparison, or scope. We refer to the tendency of such a non-gold
paragraph to be ranked as genuine evidence by the current reranker as
\emph{evidence confusability}.

For a training claim $q$, let $\mathcal{G}(q)\subseteq P$ denote its set of
gold evidence paragraphs. A base reranker assigns a relevance score
$s_0(q,p)$ to each non-empty paragraph. For each non-gold paragraph
$p\in P\setminus\mathcal{G}(q)$, we use this score as a claim-specific
and model-dependent proxy for evidence confusability, defining
$c_0(q,p)=s_0(q,p)$. A larger $c_0(q,p)$ indicates that the base
reranker ranks $p$ as a stronger competitor to gold evidence. HNR uses
only the within-paper ordering of this proxy and selects
\begin{equation}
\mathcal{N}_K(q)
=
\operatorname*{TopK}_{p\in P\setminus\mathcal{G}(q)}
c_0(q,p).
\label{eq:hard-negative-set}
\end{equation}

For each gold evidence paragraph $p^{+}\in\mathcal{G}(q)$, we form a
training group containing $p^{+}$ and the $K$ mined hard negatives
$\{p_j^{-}\}_{j=1}^{K}$ from $\mathcal{N}_K(q)$. The reranker is
optimized with a group-wise contrastive objective:
\begin{equation}
\mathcal{L}_{\mathrm{HNR}}
=
-\log
\frac{\exp s_{\theta}(q,p^{+})}
{\exp s_{\theta}(q,p^{+})
+\sum_{j=1}^{K}\exp s_{\theta}(q,p_j^{-})},
\label{eq:hnr-loss}
\end{equation}
where $s_{\theta}(q,p)$ denotes the relevance score assigned by the
fine-tuned reranker. This objective encourages gold evidence to
outrank the non-gold paragraphs ranked highest by the base reranker.
When multiple gold evidence paragraphs are available, each forms a
separate training group. \textsc{Irrelevant} instances are excluded
from HNR training because they contain no gold evidence.

At inference, HNR scores all non-empty paragraphs and ranks them in
descending order, yielding $\widehat{E}_1(q)$ for DAC and
$\widehat{E}_3(q)$ for evidence output.

\subsection{Distribution-Aligned Classification}
\label{subsec:dac}
DAC trains the relation classifier
on evidence produced by the frozen HNR. We describe its
retrieval-conditioned training construction and label-aware resampling
strategy below.

\subsubsection{Retrieval-Conditioned Classifier Training}
\label{subsubsec:retrieval-conditioned-training}

Relation classification is performed on automatically retrieved evidence
at inference. A classifier trained only on gold paragraphs, however,
observes oracle contexts that are cleaner than those available in the
complete pipeline. We therefore construct its training inputs from the
outputs of the frozen HNR, aligning the retrieval process used to form
classifier contexts across training and inference.

Specifically, after HNR training, we freeze the reranker and apply it
to the cited paper of every training claim. Let
$\widehat{E}_1(q_i)$ denote the highest-ranked paragraph for claim
$q_i$. DAC is trained on the retrieval-conditioned input
\[
x_i = [q_i;\widehat{E}_1(q_i)],
\]
with the corresponding relation label $y_i$. This construction is applied to
all four classes. In particular, Irrelevant instances also receive a non-empty
HNR top paragraph, preventing the classifier from using the absence of
evidence as a trivial label cue, an input pattern that never occurs at
inference, where every claim yields a non-empty top-ranked paragraph.

DAC is optimized through supervised fine-tuning with standard token-level
negative log-likelihood over the target relation label. Let
$\mathcal{D}'$ denote the resampled training set. The model is instructed
to generate exactly one of the four relation labels, while evidence
identifiers remain determined by HNR rather than generated by DAC.

At inference, DAC receives the same input structure,
$[q;\widehat{E}_1(q)]$, and predicts the claim--paper relation. Here,
distribution alignment refers to matching the retrieval process used to
construct classifier inputs across training and inference, rather than
assuming that their empirical distributions are identical.

\subsubsection{Label-Aware Resampling}
\label{subsec:resampling}
The four-way classification data are imbalanced: as shown in
Table~\ref{tab:resampling-analysis}, \textsc{Supported} accounts for nearly half
of the original training set, whereas each remaining class contributes
approximately 17\%. We therefore apply sampling multipliers of $1$, $2$,
$3$, and $3$ to \textsc{Supported}, \textsc{Irrelevant},
\textsc{Topical Match}, and \textsc{Overstate}, respectively. This resampling changes only the frequency of existing training instances; the development and test distributions remain unchanged.

\section{Experimental Results}

\subsection{Experimental Setup}
\label{sec:experimental-setup}

\subsubsection{Dataset.}
We use the official training and development splits of NLPCC 2026
Task 10 Track 2. The training set contains 1,945 instances, including
926 Supported, 344 Irrelevant, 338 Topical Match, and 337 Overstate
examples. The development set contains 217 instances, with 103
Supported examples and 38 examples for each of the other three
classes. Retrieval metrics are evaluated on the 179 development
instances with non-empty gold evidence, while classification and joint
metrics are computed on all 217 instances.

\subsubsection{Metrics.}
The official evaluation uses Macro-F1 and Joint@3. Macro-F1 is the
unweighted average of the F1 scores over the four relation classes.
Joint@3 counts an instance as correct only when the relation label is
correct and at least one gold evidence paragraph appears in the
predicted top three. For Irrelevant instances, the predicted evidence
list must be empty. The overall Score is the average of Macro-F1 and
Joint@3.

For evidence-reranking analysis, we additionally report Hit@$k$ and
mean reciprocal rank (MRR). Hit@$k$ measures whether at least one gold
paragraph occurs within the top $k$ positions, while MRR is computed
from the rank of the first retrieved gold paragraph.

\subsubsection{Implementation Details.}
HNR uses Qwen3-Reranker-8B and is trained with 15 mined hard negatives
per positive paragraph. The trained reranker is then frozen to construct
DAC inputs and to produce the final top-three evidence ranking. Unless
otherwise specified, DAC uses Qwen3.5-27B with LoRA fine-tuning on HNR
Top-1 paragraphs. Class resampling expands the classifier training set
to 3,639 instances, and training uses seed 42, 2,275 optimization steps,
and an effective batch size of 8. For the classifier scaling analysis,
Qwen3.5-0.8B, 2B, 9B, and 27B are trained under the same data,
optimization, and evaluation settings. 

\begin{table*}[t]
\centering
\scriptsize
\setlength{\tabcolsep}{3.5pt}
\caption{
Development-set comparison with baseline systems and component
variants of HNR-DAC. Vanilla denotes the standard
two-stage pipeline that uses the original model without hard-negative adaptation and trains the relation classifier on gold evidence. Vanilla + HNR replaces the base reranker with HNR while retaining gold-evidence classifier training. Vanilla + DAC retains the base reranker but constructs classifier training inputs from its Top-1 outputs.
}
\label{tab:main_results}
\begin{tabular}{lccrrrrrrr}
\toprule
\multirow{2}{*}{\textbf{Method}}
& \multirow{2}{*}{\textbf{HNR}}
& \multirow{2}{*}{\textbf{DAC}}
& \multicolumn{4}{c}{\textbf{Class-wise F1}}
& \multicolumn{3}{c}{\textbf{Overall}} \\
\cmidrule(lr){4-7}
\cmidrule(lr){8-10}
&
&
&
\textbf{Sup.}
& \textbf{Over.}
& \textbf{Top.}
& \textbf{Irr.}
& \textbf{Macro-F1}
& \textbf{Joint@3}
& \textbf{Score} \\
\midrule

\multicolumn{10}{c}{\textbf{Baselines}} \\
DeepSeek-V4-Pro
& -- & --
& 67.65
& 58.95
& 6.45
& 65.75
& 49.70
& 42.86
& 46.28 \\

GPT-4o-mini
& -- & --
& 66.67
& 26.09
& 3.57
& 5.13
& 25.37
& 35.02
& 30.19 \\

GPT-5.4
& -- & --
& 61.62
& 53.33
& 0.00
& 51.72
& 41.67
& 32.72
& 37.19 \\

\midrule
\multicolumn{10}{c}{\textbf{Ours}} \\
Vanilla
& -- & --
& 97.12 & 94.59 & 74.00 & 53.85
& 79.89 & 82.03 & 80.96 \\

Vanilla + HNR
& $\checkmark$ & --
& 95.81 & 94.59 & 84.71 & 73.33
& 87.11
& 88.02
& 87.57 \\

Vanilla + DAC
& -- & $\checkmark$
& \textbf{99.04} & \textbf{96.00} & \textbf{93.33} & \textbf{97.37}
& \textbf{96.44} & 93.09 & 94.76 \\

\textbf{HNR-DAC}
& $\checkmark$ & $\checkmark$
& \textbf{99.04} & \textbf{96.00} & 92.11 & 96.00
& 95.79
& \textbf{94.47}
& \textbf{95.13} \\

\bottomrule
\end{tabular}
\end{table*}

\subsection{Main Results and Component Comparison}
\label{sec:overall-development-results}

Table~\ref{tab:main_results} compares HNR-DAC with general-purpose
LLM baselines and controlled component variants. The prompted LLM
baselines perform poorly, particularly on \textsc{Topical Match},
indicating that general semantic reasoning alone is insufficient for
fine-grained evidence verification. The Vanilla pipeline achieves a
Score of 80.96. Adding HNR improves the Score to 87.57, with notable
gains on \textsc{Topical Match} and \textsc{Irrelevant}, showing that
hard-negative training improves discrimination among confusable
within-paper paragraphs.

DAC provides the largest improvement, increasing Macro-F1 from 79.89
to 96.44 and Score to 94.76. This supports training the classifier on
retrieved rather than gold evidence. The full HNR-DAC system achieves
the highest Joint@3 of 94.47 and the highest overall Score of 95.13.
Although its Macro-F1 is slightly lower than that of Vanilla~+~DAC,
the improvement in Joint@3 shows that HNR contributes stronger evidence
identification. Overall, DAC primarily improves relation classification,
while HNR complements it by improving joint label--evidence correctness.

\subsection{HNR Analysis}
\label{sec:her-analysis}

\begin{figure*}[t]
    \centering
    \includegraphics[width=0.9\textwidth, trim=0 0 0 0]{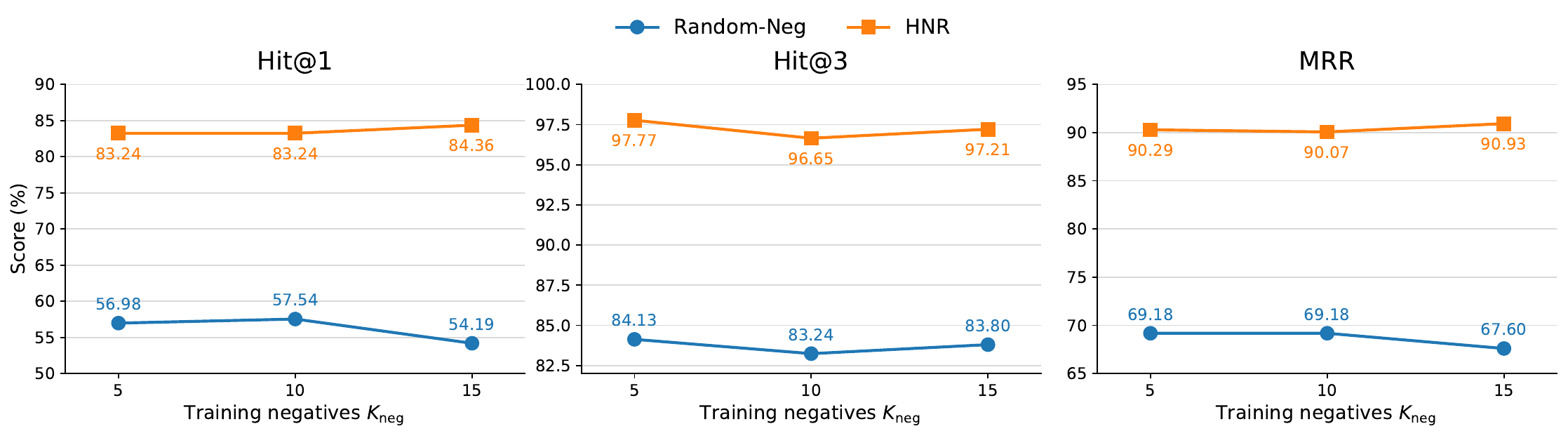}
    \caption{HNR analysis on the development set. We compare Random-Neg and HNR under different numbers of negatives per positive paragraph.}
    \label{fig:ab_HNR}
\end{figure*}

\subsubsection{Number of Hard Negatives.}

We first examine the number of mined hard negatives included with
each positive paragraph. Figure~\ref{fig:ab_HNR} reports the
results for $K\in\{5,10,15\}$. Increasing $K$ does not produce a
monotonic improvement across all retrieval metrics. HNR with $K=5$
achieves the highest Hit@3, whereas $K=15$ gives the strongest
Hit@1 and MRR. Since DAC consumes only the highest-ranked paragraph,
we select $K=15$ for the final system, prioritizing evidence quality
at the first rank rather than top-three coverage alone.

\subsubsection{Hard versus Random Negatives.}

We next compare HNR and randomly sampled negatives under the same
values of $K$. Random-Neg uniformly samples non-gold paragraphs from
the cited paper, whereas HNR selects high-scoring non-gold
paragraphs produced by the base reranker. HNR consistently
improves Hit@1 and MRR over Random-Neg for all evaluated values of
$K$, showing that high-scoring within-paper distractors provide more
informative supervision than uniformly sampled paragraphs.

The advantage is most pronounced at $K=5$, where HNR improves
Hit@1 by 26.26 points and MRR by 21.11 points. At $K=15$, it improves
Hit@1, Hit@3, and MRR by 30.17, 13.41, and 23.33 points, respectively.
These results support hard-negative mining as the primary source of
HNR's early-rank improvement.

\subsection{DAC Analysis}
\label{sec:dac-analysis}

\begin{figure*}[t]
\centering
\includegraphics[width=\textwidth]{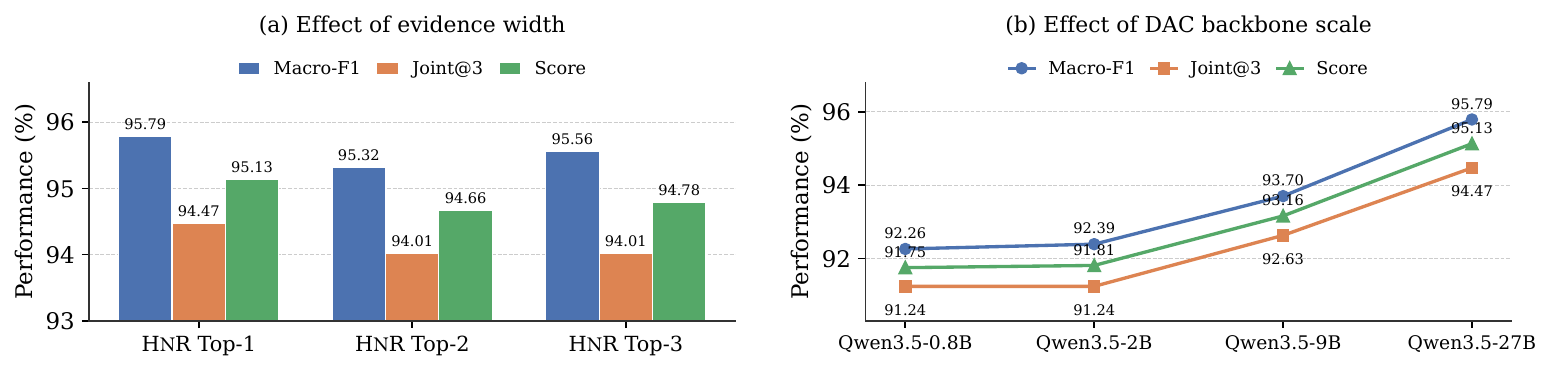}
\caption{
DAC analysis on the development set.
(a)~Effect of the number of HNR paragraphs supplied to DAC.
(b)~Effect of DAC backbone scale.
}
\label{fig:dac-analysis}
\end{figure*}

We analyze three design choices of DAC:  evidence width, classifier scale, and label-aware resampling. All configurations use the same
$K=15$ HNR and the same top-three evidence output for Joint@3, so the
reported differences arise from relation classification.

\subsubsection{Evidence Width.}

We vary the number of rank-ordered HNR paragraphs supplied to DAC,
using the same evidence width during training and inference. As shown in Figure~\ref{fig:dac-analysis}(a), Top-1 achieves the highest
Macro-F1, Joint@3, and Score. Expanding the input to Top-2 or Top-3
reduces Score by 0.47 and 0.35 points, respectively, suggesting that
lower-ranked paragraphs provide no additional benefit under the
evaluated setting. We therefore use HNR Top-1 as the DAC input while
retaining the top-three paragraph identifiers for evidence output.

\subsubsection{Effect of Model Scale.}

We further examine the effect of classifier capacity by varying the
DAC backbone from 0.8B to 27B parameters. As shown in Figure~\ref{fig:dac-analysis}(b), performance generally improves with classifier scale. The 0.8B and
2B variants obtain comparable results, while increasing the backbone
to 9B yields clearer gains in both Macro-F1 and Joint@3. The 27B
model performs best, exceeding the 9B variant by 2.09 Macro-F1 points
and 1.97 Score points. These results indicate that DAC benefits from
greater classifier capacity, although the smaller variants remain
competitive under the same retrieved-evidence interface.

\subsubsection{Label-Aware Resampling.}
\begin{table*}[t]
\caption{
Training label distributions and corresponding DAC performance before
and after label-aware resampling. Distribution cells report instance
counts with percentages in parentheses. Score is the average of
Macro-F1 and Joint@3.
}
\label{tab:resampling-analysis}
\centering
\footnotesize
\setlength{\tabcolsep}{2pt}
\renewcommand{\arraystretch}{0.5}
\resizebox{0.9\textwidth}{!}{%
\begin{tabular}{lrrrrrrrr}
\toprule
\multirow{2}{*}{Training setting}
& \multicolumn{5}{c}{\textbf{Training label distribution}}
& \multicolumn{3}{c}{\textbf{DAC performance}} \\
\cmidrule(lr){2-6}
\cmidrule(lr){7-9}
&
Sup.
& Over.
& Top.
& Irr.
& Total
& Macro-F1
& Joint@3
& Score \\
\midrule
Original
& \shortstack{926\\(47.60)}
& \shortstack{337\\(17.30)}
& \shortstack{338\\(17.40)}
& \shortstack{344\\(17.70)}
& 1945
& 95.82
& 94.47
& 95.15 \\
\shortstack[l]{Label-aware\\resampling}
& \shortstack{926\\(25.40)}
& \shortstack{1011\\(27.80)}
& \shortstack{1014\\(27.90)}
& \shortstack{688\\(18.90)}
& 3639
& 95.79
& 94.47
& 95.13 \\
\bottomrule
\end{tabular}%
}
\vspace{2pt}
\parbox{\linewidth}{\scriptsize
Sup., Over., Top., and Irr. denote Supported, Overstate,
Topical Match, and Irrelevant, respectively.
}
\end{table*}
The original DAC training set is dominated by \textsc{Supported},
which accounts for 47.60\% of the instances, whereas each of the
remaining classes contributes approximately 17\%. Since Macro-F1
assigns equal importance to all relation classes, we apply
label-aware resampling to increase the training exposure of the
minority labels.

As shown in Table~\ref{tab:resampling-analysis}, however, resampling
does not improve aggregate development-set performance. The original
distribution obtains a Macro-F1 of 95.82 and a Score of 95.15, compared
with 95.79 and 95.13 after resampling, while Joint@3 remains unchanged.
The near-identical results suggest that the 27B DAC backbone is already
able to fit the observed class imbalance on the development set.
Accordingly, we do not regard resampling as a primary source of the
development-set improvement. 

\subsection{Official Results}
\label{sec:official-results}
Table~\ref{tab:official-leaderboard} reports the official Track~2 results on the hidden test sets. Our submission ranks third overall, achieving an Average Score of 81.61, a Macro-F1 of 93.05, and a Joint@3 of 70.16. Notably, HNR-DAC obtains the highest Macro-F1 in both evaluation phases and overall, demonstrating strong generalization in fine-grained relation classification.

Compared with the development results, Macro-F1 decreases only moderately, whereas Joint@3 exhibits a substantially larger drop. This gap suggests that relation classification transfers more robustly than evidence ranking to the hidden test distribution, and that identifying annotated evidence across heterogeneous paper structures remains the primary bottleneck. Label-aware resampling was retained in the submitted system to increase exposure to minority relations; however, because no official no-resampling result is available, its independent contribution to test-set generalization cannot be isolated.

\begin{table}[H]
\caption{Official Track~2 top-three results (\%). Each block reports Score (Average Score), Macro-F1, and Joint@3; column bests are bold, second-best underlined.}
\label{tab:official-leaderboard}
\centering
\scriptsize
\setlength{\tabcolsep}{2pt}
\resizebox{0.9\columnwidth}{!}{%
\begin{tabular}{clrrr@{\hspace{7pt}}rrr@{\hspace{7pt}}rrr}
\toprule
Rank & System & \multicolumn{3}{c}{Phase~1} & \multicolumn{3}{c}{Phase~2} &
\multicolumn{3}{c}{\textbf{Overall}} \\
\cmidrule(lr){3-5}\cmidrule(lr){6-8}\cmidrule(lr){9-11}
& & Score & Macro-F1 & Joint@3 & Score & Macro-F1 & Joint@3 & Score & Macro-F1 & Joint@3 \\
\midrule
1 & System 1 & \textbf{86.21} & \underline{90.46} & \textbf{81.96} &
\textbf{86.33} & 91.89 & \textbf{80.78} & \textbf{86.27} & \underline{91.18} &
\textbf{81.37} \\
2 & System 2 & 81.28 & 86.99 & \underline{75.56} & \underline{84.70} & \underline{92.11} & \underline{77.30} &
\underline{82.99} & 89.55 & \underline{76.43} \\
3 & Our (HNR-DAC) & \underline{81.70} & \textbf{91.79} & 71.61 & 81.52 & \textbf{94.32} &
68.71 & 81.61 & \textbf{93.05} & 70.16 \\
\bottomrule
\end{tabular}%
}
\end{table}

\section{Conclusion}

We presented HNR-DAC, a two-stage framework for scientific claim
verification over cited papers. HNR improves evidence ranking through
within-paper hard negatives, while DAC trains the classifier on HNR's
top-ranked paragraph to better match inference conditions. Experiments
show that mined negatives outperform random negatives, Top-1 evidence
provides the strongest classifier input, and larger classifier
backbones further improve performance. The final system achieves a
95.13\% development Score and ranks third in the official evaluation,
with the highest overall Macro-F1. Nevertheless, the official Joint@3 results indicate that evidence-ranking generalization remains more challenging than relation classification.

\bibliographystyle{splncs04}
\bibliography{mybibliography}

\end{document}